\pdfoutput=1
\documentclass[10pt]{article}
\usepackage[margin=0.9in,top=0.75in,bottom=0.8in]{geometry}
\usepackage[T1]{fontenc}
\usepackage{mathptmx}
\usepackage{courier}
\usepackage{microtype}
\usepackage{booktabs}
\usepackage{multirow}
\usepackage{graphicx}
\usepackage{amsmath}
\usepackage{xcolor}
\usepackage{enumitem}
\usepackage[hidelinks]{hyperref}
\usepackage[numbers,sort&compress]{natbib}
\usepackage{caption}
\newcommand{\repo}{\url{https://github.com/multimodal-ai-engineering/preflight-bench}}

\begin{document}

\begin{center}
\rule{\textwidth}{1.1pt}\\[9pt]
{\fontsize{18}{23}\selectfont\bfseries What Does ``Now'' Mean to Your Computer-Use Agent?}\\[5pt]
{\fontsize{12.5}{16}\selectfont Stale screenshots, page loads, and which check to run before a click}\\[7pt]
\rule{\textwidth}{1.1pt}\\[14pt]
{\bfseries Vasundra Srinivasan}\\[2pt]
{\small Principal AI Engineer, Author---\textit{Data Engineering for Multimodal AI} (O'Reilly), Stanford School of Engineering\quad$\cdot$\quad October 2026}
\end{center}

\vspace{4pt}
\begin{center}{\large\bfseries Abstract}\end{center}
\vspace{-6pt}
\begin{list}{}{\leftmargin=0.55in\rightmargin=0.55in}\item[]
Computer-use agents plan on a screenshot for several seconds and then click at a position on the screen; if the page has changed in the meantime, the click lands on an unintended element. Prior work treats the planning interval as the source of this risk. Measuring planned clicks on 31 public web pages, we find that the risk lies in page loading: 0.4\% of clicks planned on fully loaded pages became stale within 30 seconds, against 17\% of clicks planned in the first second after a page opened. On these pages, waiting about one second for network activity to stop removes most of this risk, yet the wait rules of eight of ten widely used open-source agents leave more than 1\% of clicks stale. Commercial pages keep loading for longer: in the HTTP Archive's public crawl, 15\% of 65 desktop booking and buying home pages went 500\,ms without network requests within 3 seconds of their HTML being parsed, against 63\% of 3,028 .gov pages, and 52\% shifted their layout within 6.5 seconds of a network-idle wait capped at 3 seconds, against 8\% (61\% against 26\% among pages recorded in full). Where a check before the click is needed, as before irreversible actions and on pages that are still loading, a pixel comparison verifies the target's position as accurately as a frontier vision--language model at one eighth of the latency, and a 4B model matches the frontier model in verifying whether the previous action succeeded. We release a benchmark generator, a break-even formula for when a check pays, and three rules for agent developers: \repo.
\end{list}

\vspace{2pt}

\section{Introduction}

\begin{figure}[!b]
\centering
\includegraphics[width=\textwidth]{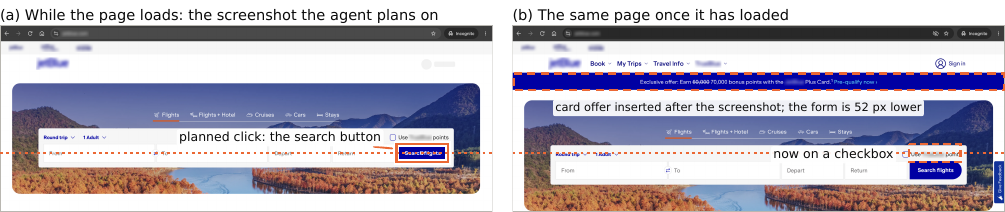}
\caption{One airline's home page in a desktop Chrome Incognito window, in screenshots we took by hand on 28 October 2026; brand names are blurred. (a)~A screenshot taken while the page loads: the search form is complete, and an agent planning on it would click the search button (box and dot; the dotted line marks its height). (b)~The same page once it has loaded: a credit-card offer was inserted under the menu and pushed the form down 52\,px, so the planned point lies on a checkbox that switches the search to loyalty points. The HTTP Archive's crawl of this page on 1~October recorded the timing: a network-idle wait capped at 3\,s ends at 3.5\,s, the layout shifts 1.3\,s later, and the network goes quiet only at 8.1\,s. Rates across sites are in Section~\ref{sec:pages}.}
\label{fig:hero}
\end{figure}

Consider an agent searching for a flight. It opens an airline's home page, waits until network activity stops or three seconds pass, captures a screenshot, and spends 6.5 seconds deciding where to click. Figure~\ref{fig:hero} shows what can happen on a real airline home page. The screenshot shows a finished search form, and the agent picks the search button. Before the click is sent, the page inserts a credit-card offer above the form and pushes the form down 52 pixels, so the click lands on a checkbox that switches the search to loyalty points. The agent's decision was correct for the screenshot it planned on; the page it acted on had changed.

Most computer-use agents follow this pattern: capture a screenshot, plan with a large multimodal model, and click at screen coordinates \citep{pusv,aoi,opencua}. Planning takes 1--13\,s \citep{pusv,aoi}. Browser-testing frameworks guard against this failure by verifying, immediately before each click, that the target element is present, stable and able to receive the click \citep{playwright}. Agents that click at coordinates omit this verification. How often the omission causes errors on real pages, and which form of perception should perform the verification, had not been measured.

\textbf{The prevailing assumption and our research questions.} Existing work treats the planning delay as the source of risk. PUSV measured a 6.5-second interval between screenshot and click and protects it with pixel re-verification \citep{pusv}; AOI and pre-compiled policy trees aim to shorten the interval \citep{aoi,aapt}. Under this assumption, the natural remedy is a fast visual check immediately before each click, which we call a \emph{preflight}. We tested the assumption and the remedy through three research questions. Throughout, a planned click is \emph{stale} when, at the moment of the click, its target has moved, disappeared or been covered.

\begin{description}[leftmargin=1.2em,itemsep=1pt,topsep=2pt,font=\normalfont\bfseries]
\item[RQ1. When does the screenshot become stale?] During the planning interval, or at another point?
\item[RQ2. Which form of perception should verify the screen?] Pixel comparison, a small vision--language model (VLM), or a frontier model?
\item[RQ3. When is a check worth its cost?] Each check adds latency to every step and occasionally raises false alarms.
\end{description}

\textbf{Findings.} \emph{RQ1:} the prevailing assumption does not hold, and planning time has little effect. On fully loaded pages, 0.4\% of planned clicks became stale within 30\,s. The risk is concentrated in the first second after a page opens, when 17\% of planned clicks become stale because the page is still loading. On these pages, waiting until network activity stops removes most of this risk, and most widely used agents do not wait long enough. Commercial pages keep loading for longer: in a public web crawl, 85\% of desktop booking and buying home pages had not reached network idle within 3\,s of DOM-ready, and half shifted their layout after the wait. \emph{RQ2:} where a check remains necessary (pages that are still loading, live dashboards, maps, shared documents and irreversible actions), the form of perception should match the question. Whether the target is still in place is a question about a few hundred pixels at one location; direct pixel comparison answers it as accurately as the best model tested, with fewer false alarms and a fraction of the latency. Whether the previous action succeeded is a question about what the screen now reports, and a 4B model answers it as accurately as the frontier model. \emph{RQ3:} a break-even formula based only on the planner's latency and the check's latency and accuracy predicted when a check pays, agreeing with held-out data on 95\% of decisions.

\begin{figure}[t]
\centering
\includegraphics[width=0.82\textwidth]{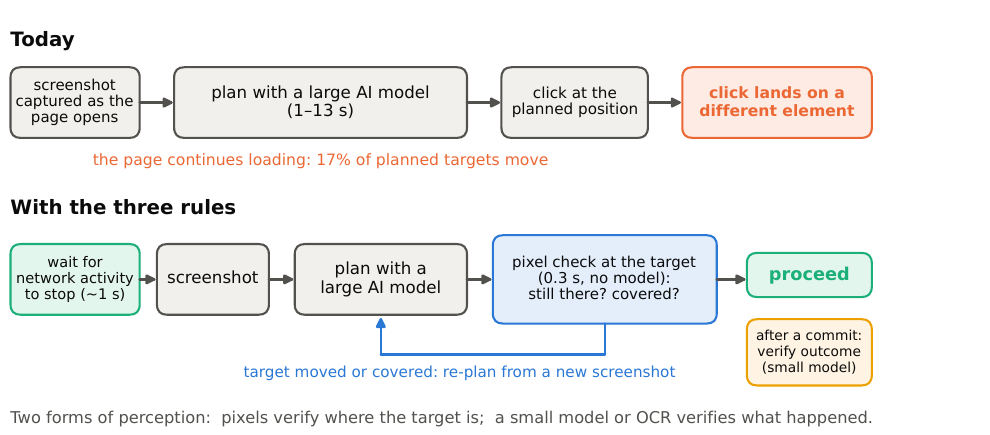}
\caption{One agent step under current practice (top) and under the three rules (bottom). Most stale clicks arise when the screenshot is captured while the page is still loading. Before the click, a pixel check confirms that the target remains where it was planned; after an action that commits a change, a small model or optical character recognition (OCR) confirms that it succeeded.}
\label{fig:diagram}
\end{figure}

\textbf{Implications.} The findings yield three low-cost rules (Section~\ref{sec:rules}): wait up to about 3\,s for network activity to stop before capturing the screenshot; before an irreversible click, verify the target with a pixel check (0.3\,s, no model), which matters more on commercial pages, since they are often still loading; and when the check fails, re-plan from a new screenshot instead of adjusting the original click. Figure~\ref{fig:diagram} in Section~\ref{sec:rules} contrasts one agent step under current practice and under these rules.

\textbf{Contributions.} (1) The first measurement of how often an agent's planned click becomes stale on real web pages, locating the risk in page loading, together with an evaluation of the post-navigation wait rules of ten open-source agents and a crawl-based measurement of how long commercial pages keep loading. (2) A direct comparison of pixel methods, OCR, small VLMs and frontier models on the three questions a pre-click check must answer. (3) A break-even formula for when a check pays, validated on held-out data. (4) An open benchmark generator that renders moving web pages with exact timing and labels every example from the page's own state. All experiments were preregistered, and we report every outcome, including refuted hypotheses.

\section{Three questions, and the checks compared}
\label{sec:checks}

A preflight runs immediately before the agent clicks and answers one of three questions: \emph{(1) Is the target still where it was planned?} \emph{(2) Is a dialog or pop-up obstructing it?} \emph{(3) Did the previous action succeed?} The first two concern a small region of pixels at one location. The third concerns content: whether the screen now reads ``Registered'' or ``Payment failed.'' This difference is why no single form of perception suits all three.

\textbf{Pixel check.} The check retains two screenshots captured 200\,ms apart at planning time. Pixels that differ between them, such as spinners, blinking cursors and tickers, change independently of the agent and are excluded. The check then searches the current screen for the target's planned image patch within 40 pixels of its original position and requires the best match to exceed any lookalike by a clear margin. Two further tests determine whether an element now covers the target and whether a dialog has appeared. The output is \emph{proceed} or \emph{re-plan}. The check takes 312\,ms, of which 200\,ms is the interval between the two screenshots; no model is involved.

\textbf{Model checks.} We posed the same questions to small open VLMs (Qwen3-VL 2B and 4B, SmolVLM2 2.2B), an OCR model (Florence-2) and two hosted models (Claude Haiku 5.5 and Claude Opus 5.5). Each model received a crop of the planning screenshot around the target, the full current screenshot and the question; for the third question, it received the screenshots before and after the action and a one-sentence description of the expected effect.

\textbf{PUSV, a published security check.} PUSV \citep{pusv} protects agents from adversaries who alter the screen deliberately. It compares the target region and the full screen with the planning screenshot and halts on any change, which is appropriate under an adversarial threat model. We reimplemented it from the paper, both as published and tuned to a 2\% false-alarm rate, to evaluate how a security check performs on ordinary pages.

\section{When is a check worth its cost?}
\label{sec:math}

A check adds latency to every step and saves time only when it intercepts a click that would have been stale. Let $C$ denote the agent's planning time, $\delta$ the check's latency, $r$ the share of stale clicks it detects (recall), $f$ its false-alarm rate, and $k$ the cost of a stale click measured in planning cycles ($k=2$ for an ordinary click, covering the wasted plan and the recovery; higher for a payment or a deletion). If $p$ is the probability that the target changes before the click, a check pays when
\begin{equation}
p \;>\; p^{*} = \frac{\delta/C+f}{r(k-1)+f}.
\label{eq:breakeven}
\end{equation}
Three consequences follow. A faster planner raises the threshold, because a stale click then wastes less time. A riskier action lowers it, because each miss is more costly. A slow check rarely pays, because its own latency extends the interval it is meant to protect. The derivation and two rearrangements, giving the maximum affordable false-alarm rate and check latency, are provided in the repository.

The same reasoning selects between two checks for a single question. A more expensive check $B$ is preferable to a cheaper check $A$ when the additional problems it detects are worth more than its additional latency and false alarms. With $q$ the rate of the problem and $M$ the cost of a missed problem in seconds,
\begin{equation}
q\,M\,(r_B-r_A) \;>\; (\delta_B-\delta_A)+(1-q)(f_B-f_A)\,C .
\label{eq:rubric}
\end{equation}

\section{Experimental setup}
\label{sec:method}

\textbf{Real pages (RQ1).} We used 31 public pages approved for read-only access, including clocks, population counters, earthquake and weather maps, reference sites and status pages. On each visit we selected up to 20 visible links and buttons and tracked, for 30\,s, whether a click at each element's planned position would still reach it. Tracking began from three states: a fully loaded page, immediately after a scroll, and immediately after the page opened, at the moment the browser has parsed the page's HTML (\emph{DOM-ready}), which usually precedes the arrival of images, fonts and data. Three rounds at different times of day yielded about 1,300 tracked targets per state. A second experiment captured the planning screenshot under seven wait rules after the page opened. A third reimplemented the post-navigation wait rule of ten open-source agents, as written in their current code, and applied it to the same pages; the agents themselves were not executed.

\textbf{Commercial pages (RQ1).} Commercial pages, where agents book and buy, usually prohibit automated access in their terms of use, so we measured them without visiting any, from the HTTP Archive's public crawl \citep{httparchive}. Its October 2026 crawl loaded 11.7 million desktop home pages once each in WebPageTest \citep{webpagetest} (desktop Chrome, cable connection, empty cache) and recorded the timing of every layout shift and network request. Among home pages ranked in the top million by the Chrome UX Report, we compared the home pages of 65 booking and buying sites from a list of 69 in six categories, from airlines to retail, written before any data was read, 3,028 .gov home pages and a sample of 4,972 home pages of sites that use e-commerce software. On each page's recorded timeline we placed the screenshot by five preregistered wait rules and an uncapped network-idle wait, and counted the page as moving if its layout shifted by more than 0.0001 in total within the next 6.5\,s. On the 31 real pages, 23\% of tracked clicks were stale in visits with such a shift after the screenshot and 0.9\% in visits without one; the share of moving pages is the primary outcome, and stale-click rates derived from it are a proxy.

\textbf{Benchmark (RQ2, RQ3).} A check remains necessary on pages that change while the agent plans, such as live dashboards, maps and shared documents, and before irreversible clicks. Real pages change too rarely to measure a check's accuracy, so we built a benchmark of six synthetic pages that move by design: a static form, a market dashboard, a news feed, a shared document edited by a second user, a panning map and a drawing board. Pages render in Chromium with a paused clock, so the interval between planning and clicking is exact (6.5\,s, the interval PUSV measured \citep{pusv}). Within that interval the generator introduces one event: no change, target moved, lookalike moved, target covered, target removed, dialog shown, or slow load. Ground truth comes from the page's element tree at the moment of the click; the checks receive only pixels. Each page has four layouts, two for tuning and two held out for testing. A third tier replays the screens under remote-desktop conditions (latency, compression and scaling). The benchmark contains 4,608 click pairs and 3,600 action-outcome pairs, and all thresholds were frozen before test labels were examined.

\textbf{Live loop.} A Claude Haiku 5.5 planner clicked named elements on the moving test pages for 400 steps per condition, with and without the pixel check.

\section{Results}

\subsection{RQ1: screenshots become stale while the page loads}
\label{sec:pages}

\begin{figure}[t]
\centering
\includegraphics[width=0.48\textwidth]{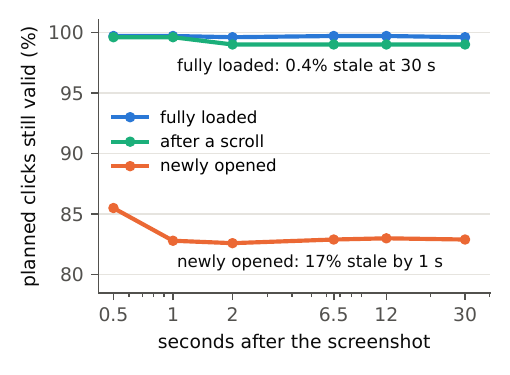}\hfill
\includegraphics[width=0.48\textwidth]{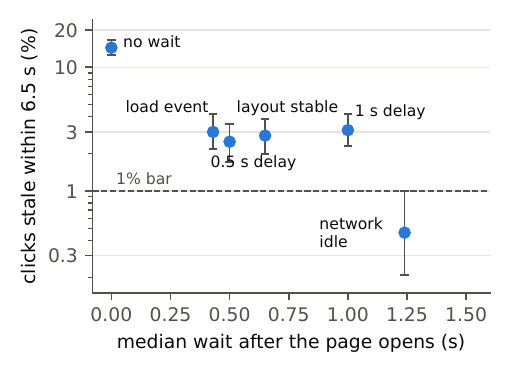}
\caption{Left: share of planned clicks that remain valid on 31 public pages, by page state. Right: median wait of each rule after the page opens, and the share of clicks it leaves stale (95\% CI). Only waiting for network activity to stop reaches 1\%.}
\label{fig:survival}
\end{figure}

Figure~\ref{fig:survival} (left) shows how long a planned click remains valid. On fully loaded pages, 0.4\% of planned clicks were stale at every interval from 0.5 to 30\,s, and the share did not increase over time. After a scroll, the share was 1.0\%. Immediately after a page opened, 14.5\% were stale by 0.5\,s and 17\% by 1\,s, after which the share stopped increasing: the page completes its layout in a burst that ends within about one second. The difference between newly opened and fully loaded pages was 17 percentage points (95\% CI 9.9--24.7), ranging from 5 points on dashboards to 38 points on a reference site.

\textbf{How long to wait.} Table~\ref{tab:settle} (left) compares wait rules. Only one fell below 1\%: waiting until the page has issued no network requests for 500\,ms (Playwright's \emph{networkidle}), which left 0.46\% of clicks stale at a median wait of 1.24\,s (90th percentile 3.2\,s). Fixed delays of 0.5 or 1\,s, the browser's \emph{load} event, and waiting for layout shifts to stop all left 2.5--3.1\%. The load event fires before much of the content arrives, and the browser's layout-shift signal does not register content that fills a fixed-size container or overlays a target, as occurs on maps.

\begin{table}[t]
\centering\small\setlength{\tabcolsep}{5pt}
\caption{Left: wait rules for the planning screenshot after a page opens, timed from DOM-ready. Right: post-navigation wait rules of ten open-source agents (code reviewed October 2026, default settings), reimplemented in our harness and applied to the same 31 pages; the agents themselves were not executed. Stale: share of planned clicks whose target moved, disappeared or was covered within 6.5\,s.}
\label{tab:settle}
\begin{tabular}{lrr@{\hskip 12pt}lrr}
\toprule
Wait rule & Wait & Stale (95\% CI) & Agent (wait rule) & Stale (95\% CI) & Wait \\
\midrule
no wait & 0\,s & 14.4\% (12.6--16.5) & Anthropic CU demo (2\,s delay) & 2.0\% (1.3--2.9) & 2.02\,s \\
0.5\,s delay & 0.50 & 2.5\% (1.7--3.5) & browser-use (load or idle) & 4.0\% (3.1--5.2) & 0.94 \\
load event & 0.43 & 3.0\% (2.2--4.2) & OpenAI CUA sample (no wait) & 15.4\% (13.5--17.4) & 0.35 \\
layout stable 0.5\,s & 0.65 & 2.8\% (2.0--3.8) & Stagehand v4 (no wait) & 12.5\% (10.8--14.5) & 0.34 \\
boxes stable 0.5\,s & 0.81 & 1.8\% (1.2--2.7) & Skyvern (load, 1\,s, animations) & \textbf{0.46\%} (0.21--1.0) & 2.03 \\
1\,s delay & 1.00 & 3.1\% (2.3--4.2) & Agent S3 (1\,s delay) & 11.7\% (10.1--13.6) & 1.02 \\
network idle & 1.24 & \textbf{0.46\%} (0.21--1.0) & OpenHands (load or idle) & 2.6\% (1.9--3.7) & 0.93 \\
 & & & Playwright MCP (load) & 3.0\% (2.2--4.1) & 0.88 \\
 & & & UI-TARS-desktop (1\,s delay) & 9.1\% (7.6--10.8) & 1.01 \\
 & & & magentic-ui (3\,s delay, then load) & 1.2\% (0.7--1.9) & 3.34 \\
\bottomrule
\end{tabular}
\end{table}

\textbf{Current agent practice.} Table~\ref{tab:settle} (right) reports the ten agents' wait rules. They leave between 0.46\% and 15.4\% of planned clicks stale, and eight of ten exceed 1\%, as preregistered. The two agents that capture the screenshot as soon as the HTML is parsed leave 12--15\%. The two that wait a fixed second leave 9--12\%: their delay starts when the server first responds, so 7\% of their screenshots are captured before the HTML has been parsed. Only Skyvern, which waits for the load event, then about one second, then for animations to finish, matches the network-idle rule. Some systems wait longer after their own actions, such as AOI's 2\,s \citep{aoi}, so this result applies to page navigation.

\begin{figure}[t]
\centering
\includegraphics[width=\textwidth]{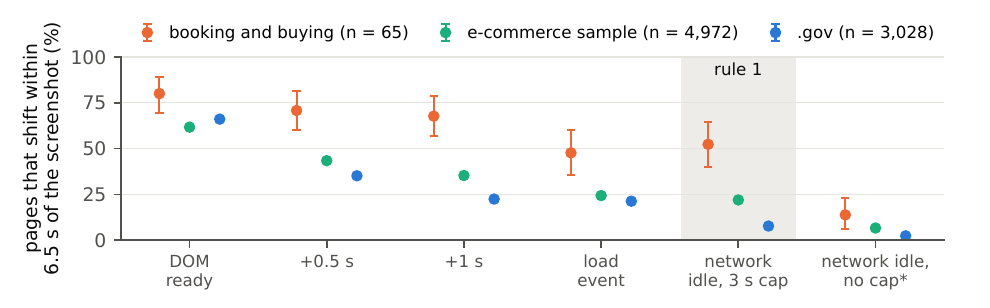}
\caption{Desktop home pages in the HTTP Archive crawl (October 2026) whose layout shifted within 6.5\,s of the screenshot, by wait rule (95\% CI). Unrecorded time counts as no shift, so each point is a lower bound; most .gov windows, and *most windows without the cap, are not recorded in full.}
\label{fig:commercial}
\end{figure}

\begin{table}[t]
\centering\small\setlength{\tabcolsep}{5pt}
\caption{Desktop home pages under rule~1 (network idle, 3\,s cap). Quiet: 500\,ms without requests within 3\,s of DOM-ready. Time to quiet (uncapped) and last layout shift: medians from DOM-ready. Shifted: within 6.5\,s of the screenshot (95\% CI); in brackets, among fully recorded pages where enough remain. $^\dagger$Post hoc.}
\label{tab:commercial}
\footnotesize
\begin{tabular}{lrrrrl}
\toprule
Home pages & Pages & Quiet within 3\,s & Time to quiet & Last shift & Shifted after the wait \\
\midrule
booking and buying & 65 & 15\% & 7.4\,s & 5.1\,s & \textbf{52\%} (40--65) [61\%] \\
\quad main host only$^\dagger$ & 57 & 11\% & 7.8\,s & 5.8\,s & 60\% (47--72) [64\%] \\
e-commerce sample & 4,972 & 19\% & 6.4\,s & 2.2\,s & 22\% (21--23) [30\%] \\
.gov & 3,028 & 63\% & 2.1\,s & 0.65\,s & 7.8\% (6.8--8.7) [26\%] \\
study sites & 21 & 71\% & 1.9\,s & 0.45\,s & 4.8\% (0--14) \\
\bottomrule
\end{tabular}
\end{table}

\textbf{Commercial pages keep loading.} After a network-idle wait capped at 3\,s (Figure~\ref{fig:commercial}, Table~\ref{tab:commercial}), 52.3\% of booking and buying pages shifted within 6.5\,s, against 7.8\% of .gov pages: a difference of 44.5 points (95\% CI 32.4--56.9; 43.4 with .gov reweighted to the booking sites' popularity ranks). Booking pages rarely go quiet within 3\,s: 15.4\% reached 500\,ms without network activity within 3\,s of DOM-ready, against 63.4\% of .gov pages, a difference of 48.1 points (38.5--56.4). Both differences held as preregistered. Eight booking pages are help, careers or partner pages on subdomains, because the crawl held no home page on those sites' main host; without them (post hoc), 60\% shifted and 11\% went quiet within 3\,s. The e-commerce sample lies between the two, desktop inner pages showed the same pattern (39\% against 5\%; exploratory), and our study sites resemble the .gov pages, consistent with the network-idle wait (median 1.24\,s) sufficing on them.

Three qualifications apply. WebPageTest stops recording once a page's network is quiet, and unrecorded time counts as no shift; for 91\% of .gov pages the window after the screenshot was not recorded in full. Among fully recorded pages the difference still held (61\% against 26\%; 35.1 points, 19.8--50.1). On mobile home pages, where 41\% of booking pages went quiet within 3\,s, the gap shrank to 12.9 points and vanished among fully recorded pages (27\% against 28\%), so the result applies to desktop. Finally, across 20 of the 21 study sites in the crawl, a crawl shift after DOM-ready did not track the share of clicks we measured as stale within 2\,s of DOM-ready (Spearman $\rho=-0.12$, $-0.56$ to $0.32$), refuting the preregistered transfer check. Stale-click rates estimated from the shift shares are therefore an unvalidated proxy, which on the study sites ran up to 4.3 times above the measured rate. For booking pages after the capped wait it gives 8.4\% or 12.5\%, depending on the calibration; divided by 4.3 (a post hoc check), both still exceed the 1.2\% break-even point for an irreversible click (Equation~\ref{eq:breakeven}, $k=10$), while for an ordinary click the data do not decide.

\subsection{RQ2: pixels answer \emph{where}, models answer \emph{what}}
\label{sec:where}

Table~\ref{tab:questions} reports each check's own decisions on the held-out moving pages. For the first question, whether the target is still in place, the pixel check detected 82\% of stale clicks at a 4\% false-alarm rate in 0.3\,s; Opus detected 83\% at 12\% in 2.4\,s. The smaller models failed in opposite directions: Haiku reported a move on almost every pair, including valid ones, while Qwen3-VL 2B and SmolVLM2 reported no move on almost every pair. Qwen3-VL 4B detected 66\% at a 37\% false-alarm rate. On real pages, PUSV as published flagged 21\% of clicks that were still valid, because ordinary pages change continually; the pixel check flagged 0.5\%.

For the third question, whether the previous action succeeded, the ranking reverses. Qwen3-VL 4B detected every failure with no false alarms, matching Opus, and the 2B model and Florence-2's OCR performed nearly as well. Dialog detection falls between the two: the pixel detector rarely raises a false alarm but misses about a quarter of dialogs on layouts it was not tuned on, whereas the models detect nearly all of them at false-alarm rates of 7--15\%.

This division is consistent with prior findings on VLMs: their image encoders capture fine positional detail that their language output does not report \citep{blindtest}, and frontier models miss most small differences between two screenshots \citep{diffspot}. To our knowledge, this is the first direct comparison of pixel methods and VLMs for verifying an exact click position.

\begin{table}[t]
\centering\small\setlength{\tabcolsep}{6pt}
\caption{Decisions of each check on held-out moving pages: share of problems detected (recall), share of valid cases incorrectly flagged (false alarms), and median latency. Small open models ran on balanced subsets; hosted models on every pair.}
\label{tab:questions}
\begin{tabular}{llrrr}
\toprule
Question & Check & Detected & False alarms & Latency (ms) \\
\midrule
\multirow{6}{*}{\shortstack[l]{Target still in place?\\(1,064 stale clicks)}}
 & pixel check & \textbf{82\%} & \textbf{4\%} & 312 \\
 & PUSV, tuned to 2\% false alarms & 24\% & 0.3\% & 5 \\
 & Qwen3-VL 2B & 0.2\% & 0\% & 1,450 \\
 & Qwen3-VL 4B & 66\% & 37\% & 2,425 \\
 & Claude Haiku 5.5 & 98\% & 95\% & 1,276 \\
 & Claude Opus 5.5 & 83\% & 12\% & 2,366 \\
\midrule
\multirow{3}{*}{\shortstack[l]{Dialog obstructing?\\(500 dialogs)}}
 & pixel dialog detector & 72\% & 1\% & 312 \\
 & Qwen3-VL 4B & 95\% & 14\% & 2,425 \\
 & Claude Opus 5.5 & 100\% & 7\% & 2,366 \\
\midrule
\multirow{4}{*}{\shortstack[l]{Previous action failed?\\(620 failures)}}
 & pixel change score & 90\% & 0\% & 312 \\
 & Florence-2 OCR & 89\% & 0\% & 4,115 \\
 & Qwen3-VL 4B & \textbf{100\%} & \textbf{0\%} & 2,814 \\
 & Claude Opus 5.5 & 100\% & 0\% & 2,366 \\
\bottomrule
\end{tabular}
\end{table}

\subsection{RQ3: when a check pays, and how to respond when it fires}
\label{sec:math-results}

\textbf{Decisions from the formula.} We supplied Equation~\ref{eq:breakeven} with values from the tuning layouts only and compared its predictions with a replay of the held-out test set. At the four change rates we measured (fully loaded, scrolled and newly opened pages, and the benchmark), its decision to check or skip matched the replay in 94.7\% of 3,000 cases, and its errors favored checking (148 of 158). The exact break-even point was less accurate: within 0.03 of the replay in 70\% of cases, below the preregistered 80\%. The discrepancies came from slow checks. A check that takes 2.4\,s is itself an outdated observation, since the target can move after the check and before the click, a point also made in robotics research on delayed actions \citep{arli}. For an ordinary click with a 6.5\,s planner, the pixel check pays once more than about 10\% of clicks would be stale; Opus does not pay at any rate up to 50\%. Equation~\ref{eq:rubric} selected the check that was less costly on the test set in 93\% of cases.

\textbf{Live agent loop.} With a Haiku planner on the moving pages, the pixel check reduced stale clicks from 65--69\% to 9--11\%. Nearly every re-plan it triggered was warranted (581 of 606). Under the preregistered accounting, which charged each re-plan at 6.5\,s, it added 28--105\% to the time per step and exceeded the preregistered 10\% limit. Charged at the planner's measured 1.1\,s, it saved 0.2--0.3\,s per step.

\textbf{Responding to a failed check.} An apparent shortcut is to locate the target again and adjust the click. This proved unsafe in every configuration. No pixel setting kept incorrect adjustments at or below 1\% (best: 11\%). Asking Haiku to relocate the element produced an incorrect click 34\% of the time and took as long as a full re-plan. In most cases where the check fired, the target had been removed or covered; the pixel method recognized this in 260 of 267 cases, whereas the models clicked an element anyway in up to 216 cases.

\section{Three rules for agent developers}
\label{sec:rules}

\begin{enumerate}[leftmargin=1.4em,itemsep=2pt,topsep=2pt]
\item \textbf{Wait for the page to finish loading before capturing the screenshot.} After navigation, wait for network activity to stop, with a cap of about 3\,s. On the public pages we measured, this removed most stale clicks at a cost of about one second per page. On desktop booking and buying home pages the wait rarely completes: 85\% had not reached network idle by the cap, and about half shifted their layout afterwards. Without the cap, the wait ends a median of 7.4\,s after DOM-ready there, so the check in rule~2 carries more of the protection.
\item \textbf{Match the check to the question.} On a fully loaded page, skip the check for ordinary clicks; at a 0.4\% change rate it does not pay. Before an irreversible action, run the pixel check at the target: it takes 0.3\,s, requires no model, and pays once about 1.2\% of such clicks would be stale. It matters more on booking and buying sites, whose pages were often still loading at the screenshot in the crawl; their checkout pages were not measured. After an action that commits a change, verify the outcome with a small model or OCR; a frontier model offers no gain for this question.
\item \textbf{When the check fires, re-plan from a new screenshot.} Do not adjust the original click.
\end{enumerate}

\noindent The broader implication concerns modality. A frontier multimodal model is well suited to planning and to reading what a screen reports, and poorly suited to confirming that a few hundred pixels remain in place. Equation~\ref{eq:rubric} makes this choice quantitative. The same timing issue likely affects training data: pipelines that record ``the frame before the action'' by a fixed rule \citep{opencua} may capture many frames during page loading. Measuring this is a natural next study.

\section{Related work}

PUSV \citep{pusv} measured the 6.5\,s interval between screenshot and click and protects it against adversaries with pixel re-verification; Atomicity for Agents \citep{atomicity} monitors page changes in browser agents. Both halt on any change and do not measure how often ordinary pages change, which \citet{vcd} identify as an open question. AOI \citep{aoi} and pre-compiled policy trees \citep{aapt} shorten the interval by acting faster. The limitations of VLMs in fine spatial judgment are documented by BlindTest \citep{blindtest}, DiffSpot \citep{diffspot} and VisCritic \citep{viscritic}, and \citet{belcak} argue for small models in agent workflows. Robotics addresses a 40--400\,ms form of the same delay by predicting the robot's own motion \citep{rtc,vlash,a2c2,arli}; web pages offer no comparable motion model, which explains why adjusting the click fails here. Browser-testing frameworks wait for elements to stabilize before acting \citep{playwright}. Robustness benchmarks inject disruptions into agent tasks: network delays, server errors and pop-ups through a proxy (WAREX \citep{warex}), shifted layouts and pop-ups on synthetic sites (StressWeb \citep{stressweb}), interface changes between actions in desktop applications (DynamicGUIBench \citep{dynamicgui}), and deceptive pop-ups and banners in a self-hosted shop (WebDecept \citep{webdecept}). In real-user data, home pages are slightly less stable than other pages, which the Web Almanac links to hero media, banners and promotions \citep{almanac}. None of these measures how often real pages change after an agent's screenshot or tests wait rules.

\section{Limitations and preregistration}

The benchmark pages change far more often than real pages (about 70\% versus 0.4\% of planned clicks); they serve as a stress test for checks, while the real pages provide the base rate. The real-page set comprises 31 public, non-commercial pages whose terms permit automated access, in one browser over three days. Commercial pages were measured from a public crawl, without visits, because their terms of use typically prohibit automated access. The primary measurement covers desktop home pages (checkout and other pages behind a login or cart were not crawled), one throttled load per page from US test locations, and one crawl month; it counts unrecorded time as no shift, misses content that covers a target without shifting the layout (18\% of stale clicks on the study pages), and did not track stale clicks across the study sites, so it measures how long commercial pages keep loading and moving on desktop and leaves their stale-click rate unmeasured. Authenticated applications that update continuously may also behave differently. The agent results reimplement each agent's wait rule; the agents were not run end to end. The perfect outcome-detection score at 4B suggests that our failure set is relatively easy. We used a single planner, and about 70\% of its own errors stemmed from similar labels produced by our task generator, which no pre-click check can detect.

We preregistered fourteen hypotheses across five registrations, each frozen before its data was collected or read. Six were supported: the formula's check-or-skip decisions match held-out data; Equation~\ref{eq:rubric} selects the less costly check; clicks become stale more often immediately after navigation than on a loaded page; at least eight of ten agents' wait rules leave more than 1\% of clicks stale; and, on desktop home pages, booking and buying pages shift after the capped network-idle wait more often, and reach network quiet within 3\,s less often, than .gov pages. Eight were refuted, including that a model of 4B parameters or fewer would come within 2 points of Opus on every question, that the live check would add less than 10\% to step time, that the break-even point would fall within 0.03 in 80\% of cases, that a 0.5\,s layout-shift wait would reach 1\%, that model-based click adjustment would be safe, and that the crawl's shift signal would track stale clicks across the study sites. All outcomes are reported in the repository.

\section{Conclusion}

For computer-use agents, the principal risk lies in page loading more than in planning: on the public pages we measured, screenshots become stale mostly within the first second after a page opens, and waiting about one second for network activity to stop removes most of that risk. On desktop, the home pages of sites where agents book and buy keep loading past a 3-second wait, so on them the check before an irreversible click matters more. Where a check remains necessary, pixel comparison should verify where the target is, a small model should verify what happened, and a failed check should trigger a new plan.

\paragraph{Code and data.} Benchmark generator, pages, checks, crawl queries, preregistrations, raw model outputs and analysis code: \repo. Real-page screenshots are not released because they contain third-party content.

\paragraph{Disclaimer.} This is independent research conducted outside any employment, with no funding and no use of non-public information. It is not endorsed by our employer or by any vendor named here.

\bibliographystyle{unsrtnat}
\bibliography{refs}
\end{document}